\documentclass[11pt,draftclsnofoot,onecolumn]{IEEEtran}
\usepackage{lmodern}
\usepackage[hidelinks]{hyperref} 
\usepackage{fullpage}
\usepackage{graphicx}
\usepackage{epstopdf,color}
\usepackage{caption}
\usepackage{amsmath}
\usepackage[inline]{enumitem}
\usepackage{fancyhdr}
\usepackage[headsep=20pt]{geometry}

\setlist[itemize]{leftmargin=*, noitemsep, topsep=0pt, parsep=0pt, partopsep=0pt, itemjoin={\newline}}

\usepackage{array}
\usepackage{ragged2e}
\newcolumntype{P}[1]{>{\RaggedRight\hspace{0pt}}p{#1}}

\fancypagestyle{firststyle}
{
  \fancyhf{}
	\setlength{\headheight}{0.5in}
  \chead{\small \textsf{Extended version of the article V. A Mateescu and I. V. Baji\'{c}, "Visual attention retargeting," \textit{IEEE MultiMedia}, vol. 23, no. 1, pp. 82-91, Jan.-Mar. 2016.}\\ \vspace{0.5in}}
}

\title{Adversarial Attacks on Human Vision}
\author{Victor A. Mateescu and Ivan V. Baji\'{c}\\School of Engineering Science, Simon Fraser University,  Canada
}

\begin{document}

\maketitle


\begin{abstract}
This article presents an introduction to visual attention retargeting, its connection to visual saliency, the challenges associated with it, and ideas for how it can be approached. The difficulty of attention retargeting as a saliency inversion problem lies in the lack of one-to-one mapping between saliency and the image domain, in addition to the possible negative impact of saliency alterations on image aesthetics. A few approaches from recent literature to solve this challenging problem are reviewed, and several suggestions for future development are presented.
\end{abstract}

\thispagestyle{firststyle}

\section{Visual Saliency}
\label{sec:VisualSaliency}

Visual attention retargeting is a field of research in which the content of an image or video is altered in an effort to guide viewer's attention from where it normally would be, towards another, desired location.\footnote{More recently, when applied to machine vision rather than human vision, these kinds of methods have been referred to as \textit{adversarial attacks}.} The origins of attention retargeting are closely tied to visual saliency, a measure of propensity for drawing visual attention. A considerable amount of research has been done on saliency computation~\cite{Borji_Itti_PAMI_2013}; we can confidently predict the areas of an image that are likely to draw our immediate attention. The problem of going in the reverse direction, i.e., dictating where we should look and finding the image that satisfies this requirement, has been relatively unexplored. Since attention retargeting builds upon principles of early attention, we begin our overview with a discussion of visual saliency.

A full, detailed analysis of every object within view at any single point in time is an incredibly complex task, far beyond the capabilities of human visual processing~\cite{Tsotsos1990}. And yet, most would agree that the simple act of seeing is rather effortless; watching television is not often thought of as a taxing activity. To cope with the vast amount of input obtained through our eyes~\cite{Itti2001}, our attentional mechanisms will focus on a few key areas for in-depth analysis using a two-stage process~\cite{Treisman1980}.

\begin{figure}
\centering
\begin{minipage}[b]{0.38\linewidth}
  \centering
  \centerline{\includegraphics[width=4cm]{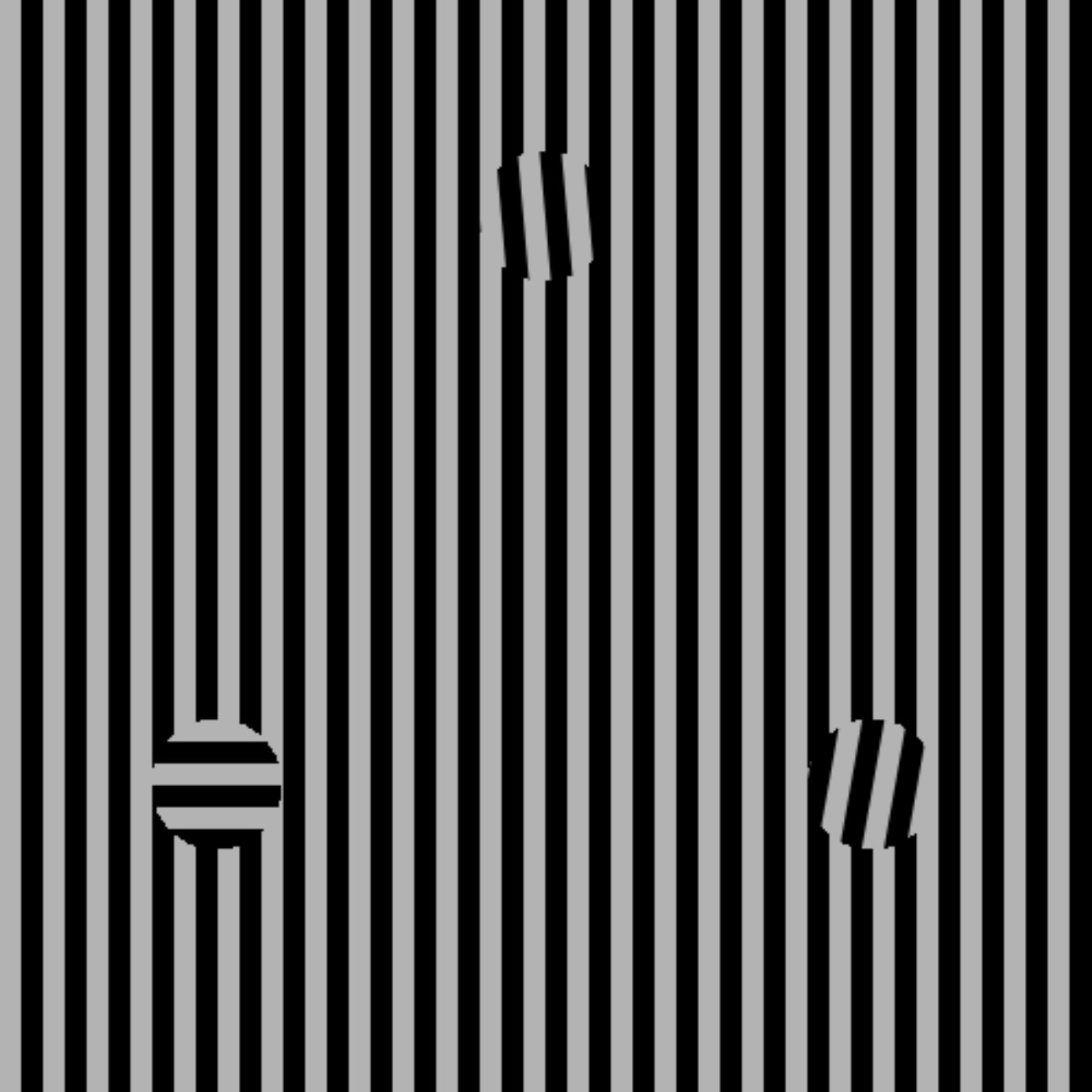}}
  \centerline{(a)}\medskip
\end{minipage}
\begin{minipage}[b]{0.38\linewidth}
  \centering
  \centerline{\includegraphics[width=4cm]{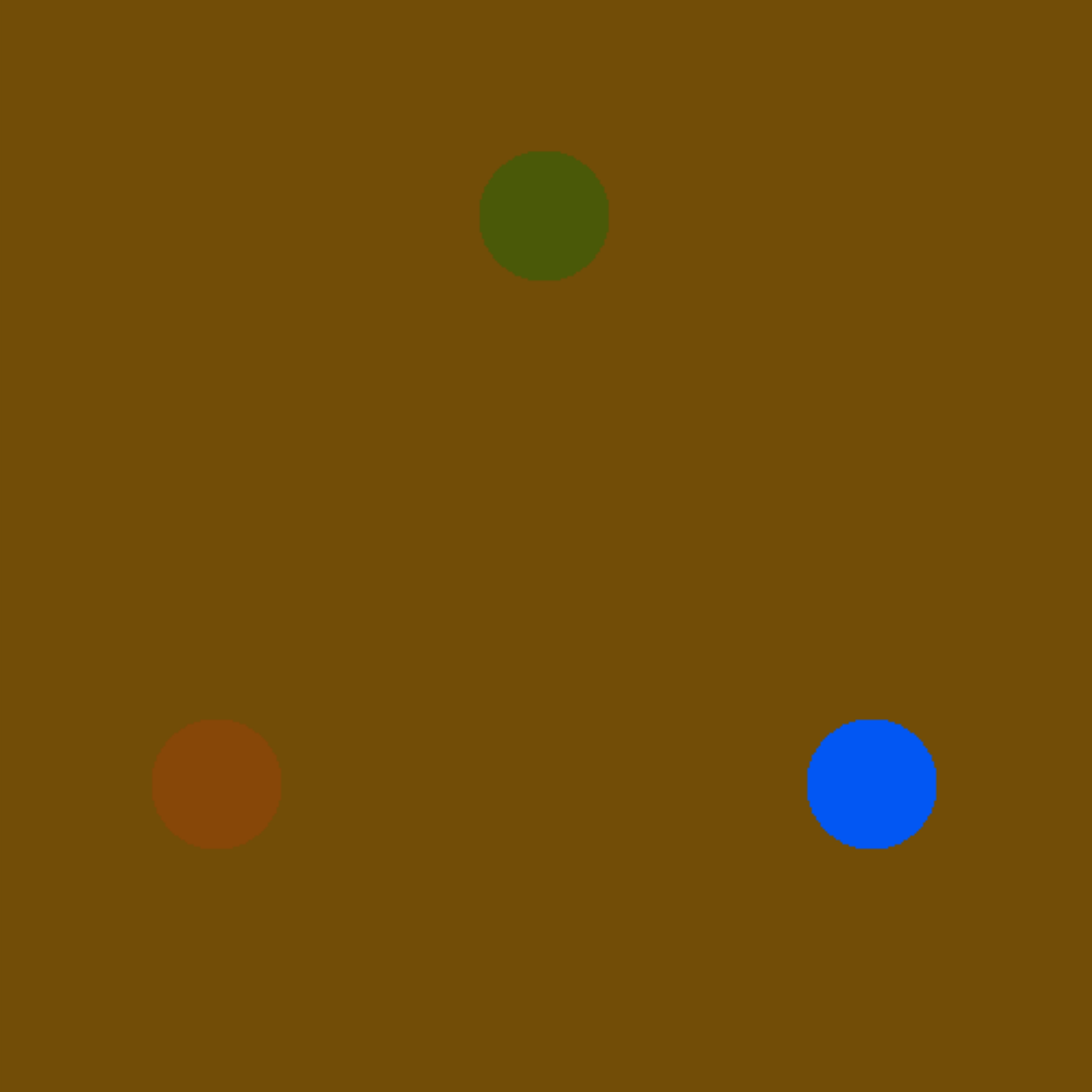}}
  \centerline{(b)}\medskip
\end{minipage}
%
\caption{Illustration of visual saliency in terms of (a) orientation and (b) color, from~\cite{Mateescu2014Color}.}
\label{fig:ColorOrientationExample}
\end{figure}

The initial processing narrows down the list of potentially relevant areas across our entire visual field~\cite{Itti2001} by identifying objects that appear to pop out in terms of basic visual features, such as color, orientation, and motion~\cite{Wolfe2004}. An interesting or conspicuous object (in the context of these visual primitives) can be perceived within 25 - 50 ms~\cite{Itti2001} in an effortless manner, regardless of clutter from uninteresting objects~\cite{Treisman1980}. The second stage of attention involves a more complex, detailed processing out of the viewer's own volition. Although the deliberate nature of this mechanism can override the attention given to the objects that popped out in the first stage, it cannot be deployed at such a fast rate. This means that if the scene suddenly changes, we involuntarily draw our gaze toward objects that are sufficiently salient in the pre-attentive stage regardless of any intent to avoid them. For example, you'll likely be distracted for a moment by the backlit mobile phone screen in a dark movie theater the instant someone in front of you takes it out to start texting. Soon afterwards however, you resolve to ignore it and return your focus onto the movie.

\begin{figure}
	\centering
	\centerline{\includegraphics[width=13cm]{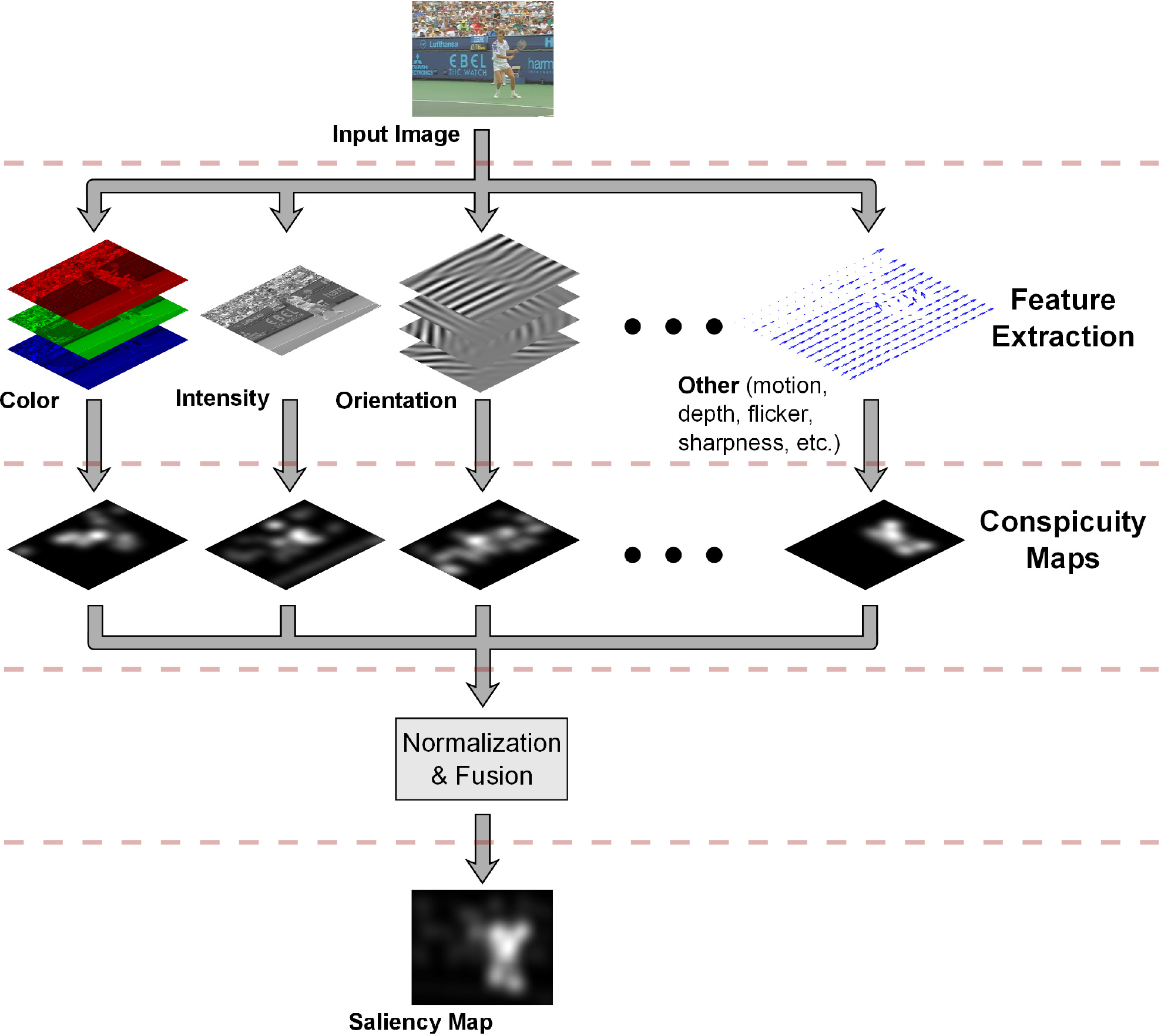}}
\captionsetup{justification=centering}
\caption{A popular model for visual saliency computation \cite{Itti1998}.}
\label{fig:SaliencyComputation}
\end{figure}

Visual saliency primarily refers to the initial stage of visual attention. A region of interest (ROI) is salient if it is perceptually different from its local surroundings in terms of basic visual features (Fig.~\ref{fig:ColorOrientationExample}). A popular architecture for visual saliency computation~\cite{Itti1998} is illustrated in Fig.~\ref{fig:SaliencyComputation}. A set of visual features that drive early attention is extracted first from the input stimuli. In the subsequent stage, conspicuities are identified within each feature channel by computing local dissimilarity among features corresponding to each region of the input. For example, image luminance can be divided into non-overlapping square patches, each of which is compared against their adjacent patches. This operation, frequently referred to as ``center-surround,'' produces a conspicuity map for each feature channel, which details the particular feature's influence on the overall saliency. Each conspicuity map is normalized to ensure that their values fall within a common range and to punish uniformity, and then combined with other maps. 

The final result is a grayscale saliency map, as illustrated in Fig.~\ref{fig:SaliencyComputation} (bottom). Highly salient regions that are predicted to draw viewer's attention are denoted by bright pixels, whereas non-salient regions that are likely to be neglected are indicated by dark pixels. The above architecture can be applied to predict saliency in videos as well. The input would be a sequence of images, and temporal features, such as flicker and motion, would be utilized in addition to spatial features to produce a saliency map for each image in the sequence.

The past few decades of research demonstrate a fairly decent grasp of predicting visual saliency~\cite{Borji_Itti_PAMI_2013}. This is significant because the involuntary nature of visual saliency makes it a powerful tool for orienting attention. We may then pose an interesting problem: \textit{if we can reasonably accurately estimate visual saliency, how can we alter it to manipulate viewers' attention}? In the following section, we describe attention retargeting as a saliency inversion problem and specify the main challenges involved. Various approaches to this problem are provided with examples from existing work. We finish our overview with a discussion of subliminal attention guiding and provide some suggestions for future work.

\section{Attention Retargeting}
\label{sec:AttentionRetargeting}

\begin{figure}
	\centering
	\centerline{\includegraphics[width=11cm]{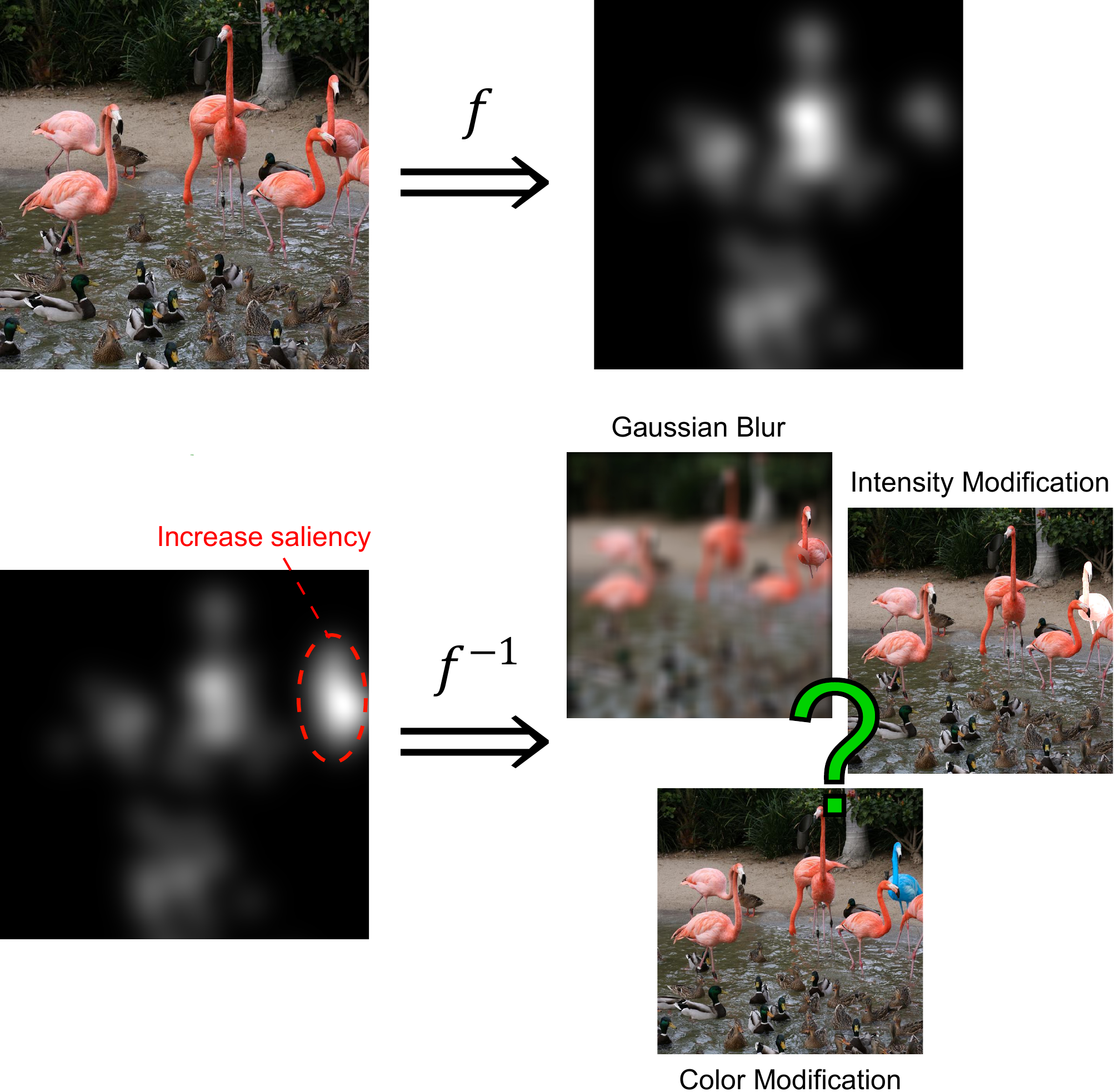}}
\captionsetup{justification=centering}
\caption{Image and corresponding saliency (top). Target saliency map and possible retargeted images (bottom).}
\label{fig:ProblemFormulation}
\end{figure}

Attention retargeting refers to modifying an image or video in an effort to alter viewer's gaze patterns in a desired way.
This can be thought of as a saliency inversion problem, as illustrated in Fig.~\ref{fig:ProblemFormulation}. Given an image and a (target) map of desired saliency, we want to obtain the modified image whose saliency matches the target saliency map. Unfortunately, this is an ill-posed problem since there is no one-to-one mapping between saliency and images. This is largely due to the fact that saliency can stem from various different features. As illustrated in Fig.~\ref{fig:ProblemFormulation}, the desired change in saliency can be achieved through the manipulation of features like intensity, color, or spatial frequency, among others. This problem can persist even when the retargeting is constrained to a single feature, since saliency is based on context. For example, the saliency of a ROI can be altered by raising the intensity of the ROI, or decreasing the intensity of its surroundings, or a combination of the two.

The fact that a desired change in saliency can be obtained in many ways leads to the impression that modifying saliency is easy. This is a difficult point to contest; a drastic change in the intensity of a region can easily make it stand out, and an application of Gaussian blurring to reduce visual conspicuities may serve to conceal it. Although modifying saliency may be simple in itself, doing so in a manner that preserves the aesthetics of the original image is not. The concept of aesthetics is fairly subjective and difficult to measure.

Attention retargeting is relatively unexplored and lacks a unified framework. In the remainder of this section, we present three broad classes of methods that have been developed for this purpose.
Examples and ideas for improvement are provided with a brief overview of existing work in the field.

\subsection{Iterative Black-Box Approach}
\label{subsec:OptApproach}

\begin{figure}
	\centering
	\centerline{\includegraphics[width=15cm]{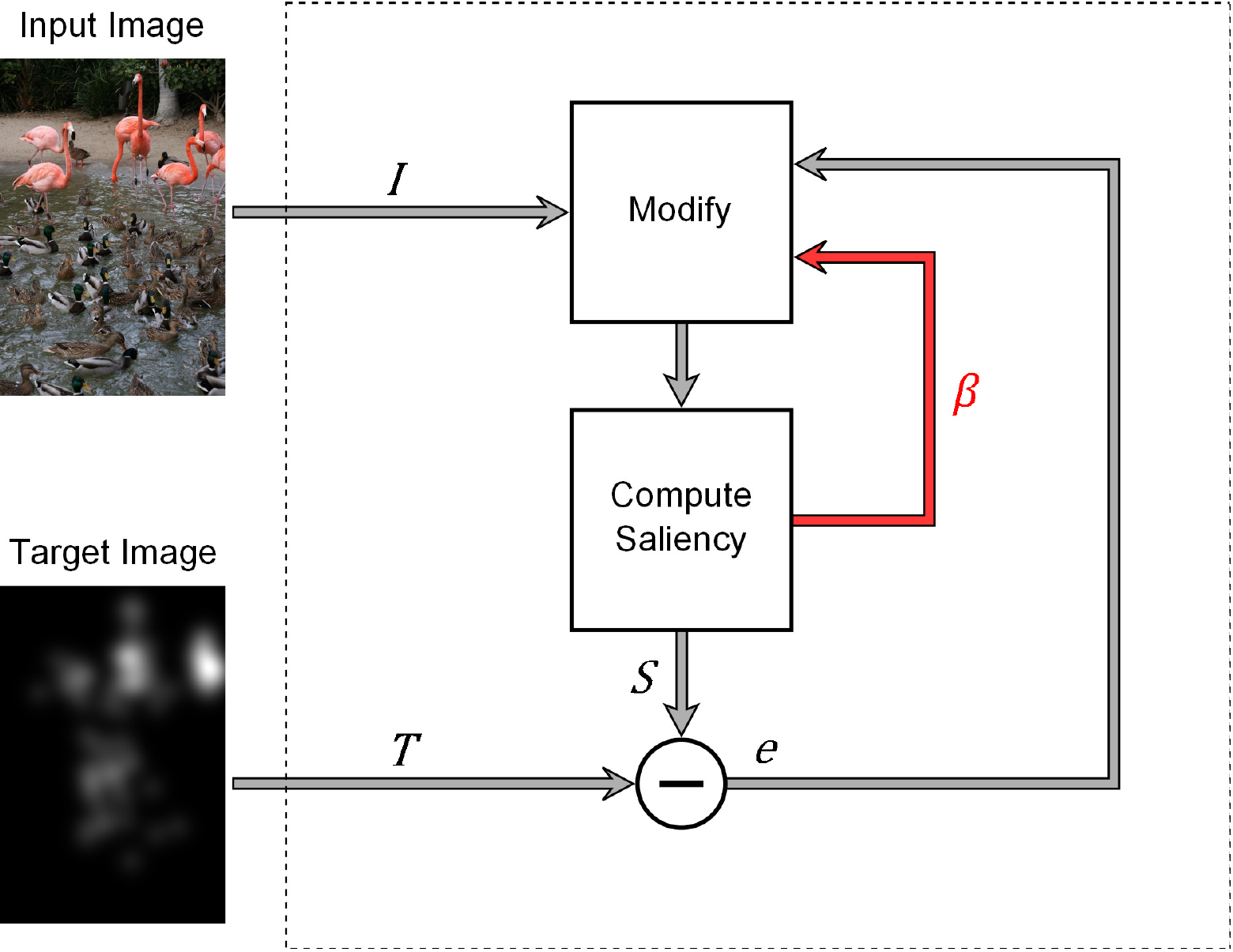}}
\captionsetup{justification=centering}
\caption{Attention Retargeting as an optimization procedure. The signal shown in red indicates that the model uses intermediate outputs of saliency computation (feature and conspicuity maps) to help determine the correct modification at each step.}
\label{fig:OptDiagram}
\end{figure}

The simplest approach to attention retargeting is to use saliency computation as a black box in an iterative optimization procedure, as shown in Fig.~\ref{fig:OptDiagram} (excluding the the signal drawn in red). At each iteration, the saliency $S$ of the image $I$ is obtained after it has been modified with respect to a chosen set of features. These features can be different from those used to compute saliency. The goal is to minimize the error between the saliency $S$ and the target saliency $T$, given by $e = T - S$.
In order to maintain the naturalness or aesthetics of the input image, a set of constraints can be introduced to prevent overmodification. The process ends when $e$ becomes small enough, or when the modifications can no longer produce an appreciable change in saliency towards the intended goal.

An example of this approach can be found in the work of Wong and Low~\cite{Wong2011}. In their model, the user segments an image into $N$ segments and enumerates them in order of importance. A target importance value $T_{i}$ is assigned to each segment, where $i = 1$ is the most important segment and $i = N$ is the least important. Modifications are made to the intensity, color saturation, and sharpness of the original image until the average saliency within each segment matches the target importance value for the corresponding segment. Specifically, the saliency error is defined as

\begin{equation}
e = \sum\limits_{i=1}^N |\mathcal{N}(T_{i}) - \mathcal{N}(S_{i})|,
\label{eq:WongLowObjectiveFunction}
\end{equation}

where 
$S_{i}$ is the average saliency of the $i$th segment, and $\mathcal{N}(\cdot)$ is a normalization operator defined as $\mathcal{N}(X_{i}) = X_{i}/\sum_{j=1}^N X_{j}$.  

The main disadvantage of this approach is the limited insight into why the images are modified the way they are. This limitation stems from the use of saliency computation as a black-box function, which only lets us observe the effect of a modification after the change to the image has been made. If $\mathcal{N}(T_{i}) - \mathcal{N}(S_{i})$ is small for some $i$, no modifications are necessary in that region. If $\mathcal{N}(T_{i}) - \mathcal{N}(S_{i})$ is large and positive then we have to increase saliency, and if $\mathcal{N}(T_{i}) - \mathcal{N}(S_{i})$ is large and negative then we have to decrease saliency in order to match the target saliency. All we can do is modify the image where $|\mathcal{N}(T_{i}) - \mathcal{N}(S_{i})|$ is large, recompute the saliency, and see whether the cost function is actually minimized. Since the modification may not necessarily alter saliency in the right direction, many iterations may be needed.

This can be remedied if the saliency computation block contains feature channels that are pertinent to the modifications being made. For example, let us consider a saliency detector that extracts spatial frequency information from blocks of $I$ to produce a feature map $\beta_{f}$. Center-surround operations are applied on $\beta_{f}$ to produce the conspicuity map $\beta_{c}$, describing the saliency of $I$ that originates from spatial frequency. The modification in this example is either blurring or sharpening (e.g., via Gaussian or Laplacian filters) applied to specific regions of $I$, which decrease or increase their spatial frequency, respectively, by varying degrees. 

Consider the case where saliency must be reduced in a particular region.
If $\beta_{c}$ is small in this region, then the feature associated with it does not contribute to saliency and does not need to be modified. Conversely, if $\beta_{c}$ is large then the feature in $\beta_{f}$ must be altered in the opposite direction. In our example, if $\beta_{f}$ indicates high spatial frequency in that region of $I$ then we need to blur it; otherwise, we need to sharpen it. Similar logic is applied in the case where saliency must be increased 
in a particular region. A large $\beta_{c}$ in this region indicates that it is already salient with respect to its corresponding feature, hence no modification is needed. If $\beta_{c}$ is small in this region, then the corresponding region of $I$ must be made salient with a modification that alters the feature in $\beta_{f}$ in the opposite direction, as before. Thus, the information extracted during saliency computation $\beta = \{\beta_{f},\beta_{c}\}$ can be used to guide the direction and possibly even the magnitude of the modifications at each step. This improved algorithm, shown by the addition of the red signal in Fig.~\ref{fig:OptDiagram}, can be thought of as a steepest 
descent algorithm for attention retargeting.

Hagiwara \textit{et al.} apply this methodology in~\cite{Hagiwara2011}. They reverse engineer a simplified version of the well-known saliency model by Itti \textit{et al.}~\cite{Itti1998}, which decomposes an RGB image into a set of four color features $r = R - (G + B)/2$, $g = G - (R + B)/2$, $b = B - (R + G)/2$, and $y = (R + G)/2 - |R - G|/2 - B$, and an additional intensity feature $v = (R + G + B)/3$. Conspicuity maps are then obtained for each feature and combined to compute saliency. Hagiwara \textit{et al.} first select a ROI whose saliency is to be maximized. They determine the change in the image channels $\Delta R$, $\Delta G$, $\Delta B$ needed to increment saliency at each pixel as a function of the features and their corresponding conspicuity maps. Since the saliency of the ROI is to be maximized relative to the rest of the image, each pixel within the ROI is modified by $\Delta \alpha$ (calculated individually for each pixel), and pixels outside are modified by their respective $-\Delta \alpha$, where $\alpha \in \{ R, G, B\}$.

\subsection{Direct Mapping}
\label{subsec:DirectMapping}

The optimization procedure in Fig.~\ref{fig:OptDiagram} uses feedback from saliency computation to estimate how the image is to be modified in a series of steps. Mapping an additive or multiplicative change to the saliency map (or conspicuity map of a particular feature) directly onto the image domain is a highly non-trivial task. However, one of the earlier works on attention retargeting by Su \textit{et al.}~\cite{Su2005}, which de-emphasizes distracting textures in images, demonstrates that this is indeed possible. Since there currently are no generalizations of this approach, we illustrate the concept of direct mapping by summarizing the methodology of Su \textit{et al.}.

\begin{figure}
	\centering
	\centerline{\includegraphics[width=15cm]{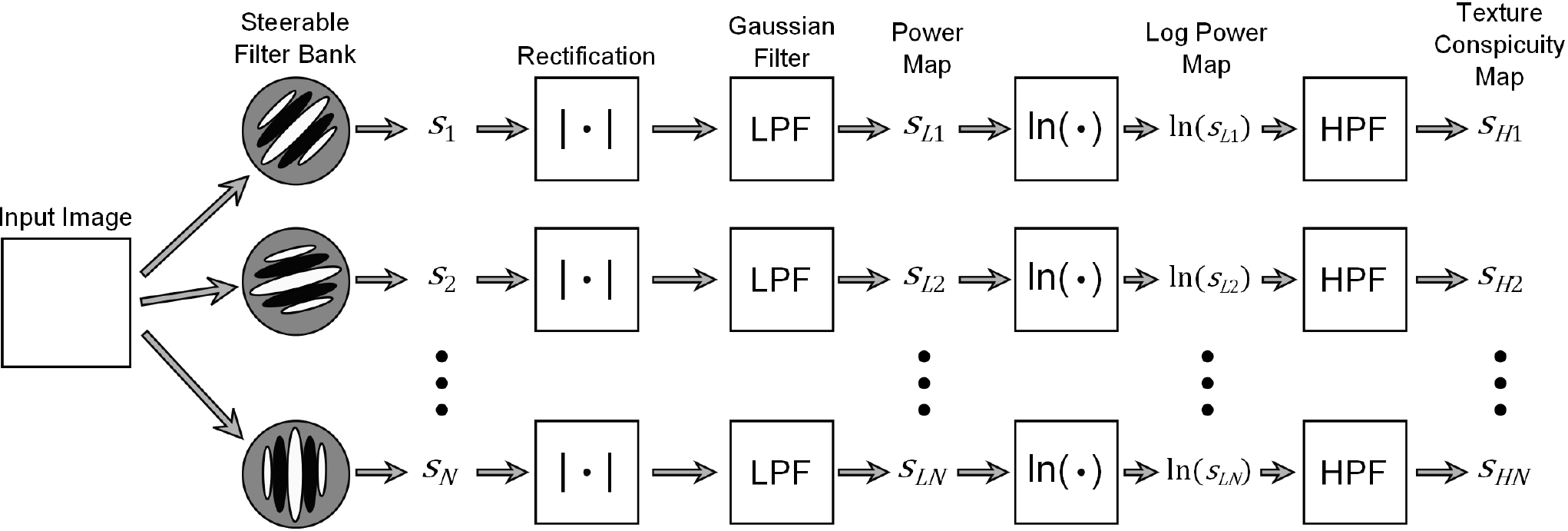}}
\captionsetup{justification=centering}
\captionsetup{justification=centering}
\caption{Computation of visual saliency from texture using steerable filter banks~\cite{Su2005}.}
\label{fig:SuDiagram}
\end{figure}

The texture-based saliency of an image is computed using steerable filter banks, as shown in Fig.~\ref{fig:SuDiagram}. The local frequency content $s_{Ln}$ is computed for each subband by rectification (since $s_{n}$ is band-limited) followed by local averaging with a Gaussian filter. Conspicuities within each subband $s_{Hn}$ can be found by applying a high pass filter to the local frequency content. Regions of the image with highly salient textures are identified by large values in the conspicuity maps $s_{Hn}$.

Minimizing the saliency of these regions can be thought of as removing any conspicuities from the local frequency content, i.e., subtracting $s_{Hn}$ from $s_{Ln}$. To ensure that the values of $s_{Ln}$ remain positive after any modification, the procedure is revised so that any subsequent processing of $s_{Ln}$ is done in the log domain. Thus, $s_{Hn}$ is subtracted from $\ln(s_{Ln})$ instead, which is equivalent to a scaling of the original coefficients
\begin{equation}
s_{n}' = s_{n}e^{-s_{Hn}}.
\label{eq:SuModification}
\end{equation}
The modified image can be reconstructed as a linear combination of the modified subbands.

\subsection{ROI-Based Retargeting}
\label{subsec:RoiBased}

Often, it is only desirable to draw attention to a specific ROI, or perhaps a few ROIs. For example, the target saliency map in the bottom-left of Fig.~\ref{fig:ProblemFormulation} indicates that we want the flamingo in the top right of the image to draw more attention than the rest of the image. To this end, we may supply a simple binary mask of the desired ROI with the objective of making this region as salient as possible relative to the rest of the image. 

Kim and Varshney apply this concept in a straightforward manner to guide attention towards selected regions in visualizations of volumetric datasets~\cite{Kim2006} and 3-D meshes~\cite{Kim2008}. They model saliency as the convolution of a center-surround kernel, e.g., difference of Gaussians, with a feature map of an image. This rudimentary saliency computation can be written as a matrix multiplication $\mathbf{Cx} = \mathbf{s}$, where $\mathbf{C}$ is the center-surround operation matrix, while $\mathbf{x}$ and $\mathbf{s}$ are, respectively, the vectorized feature and saliency maps. Suppose instead that we supply $\mathbf{s}$ as a binary mask of the ROI, with the goal of solving for an unknown $\mathbf{x}$. If $\mathbf{C}$ is well-conditioned then $\mathbf{C}^{-1}\mathbf{s}$ is a vector of scaling factors that can be used to modulate feature channels (e.g., intensity and color saturation in~\cite{Kim2006}) to increase the relative saliency of the ROI.

The absence of the target saliency as an input highlights a big redundancy---the lack of need for full-scale saliency computation---in the approaches outlined in Section~\ref{subsec:OptApproach}. Saliency computation typically requires center-surround differences over the whole image. With the simplified goal of either increasing or decreasing saliency within a single region, it suffices to perform this operation on the ROI and its surroundings alone, rather than the entire image.

\begin{figure}[t]
\begin{minipage}[b]{.48\linewidth}
  \centering
  \centerline{\includegraphics[width=9cm]{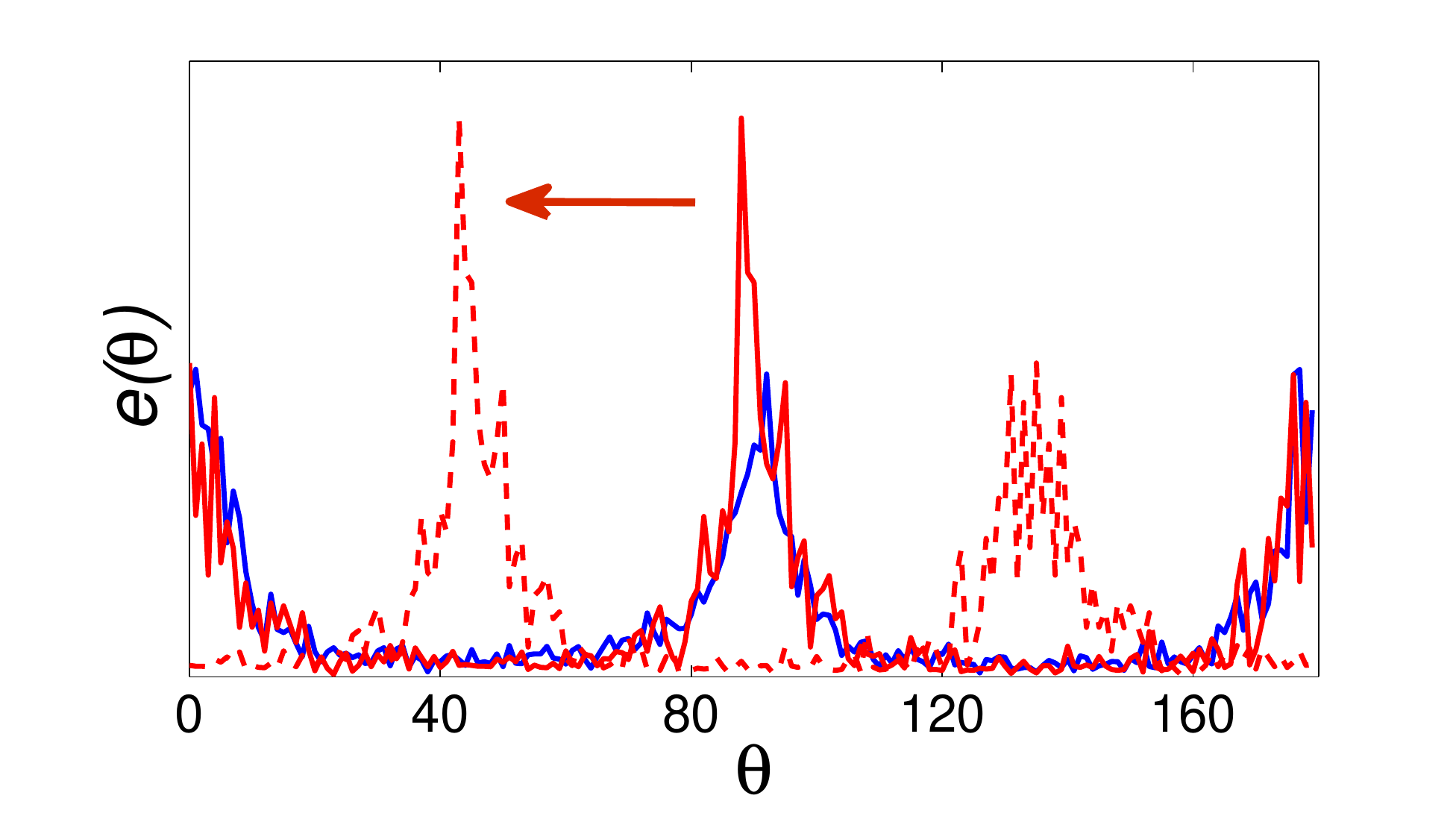}}
  \centerline{(a)}\medskip
\end{minipage}
\begin{minipage}[b]{.48\linewidth}
  \centering
  \centerline{\includegraphics[width=6.5cm]{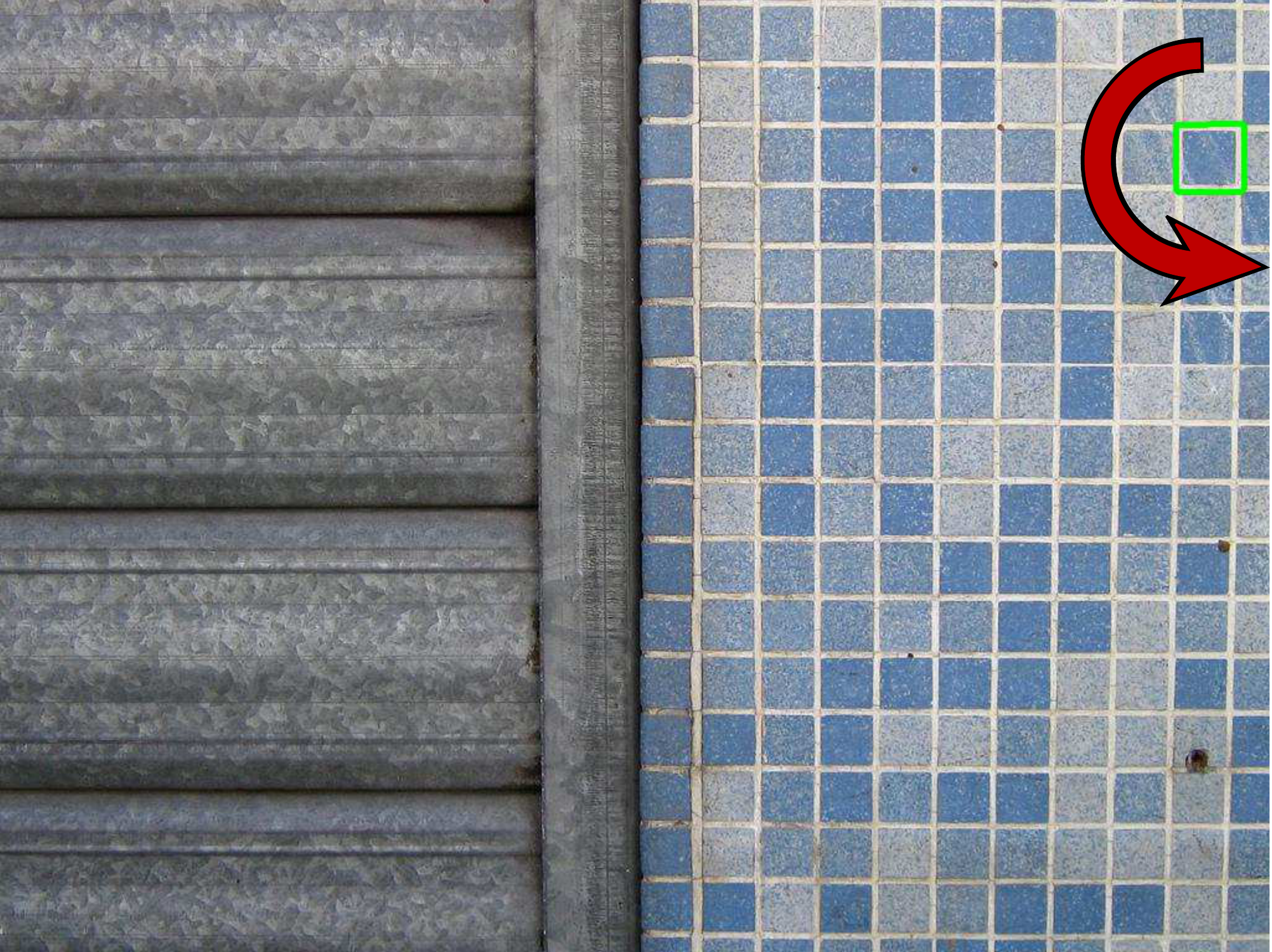}}
  \centerline{(b)}\medskip
\end{minipage}
\centering
\begin{minipage}[b]{.48\linewidth}
  \centering
  \centerline{\includegraphics[width=7.5cm]{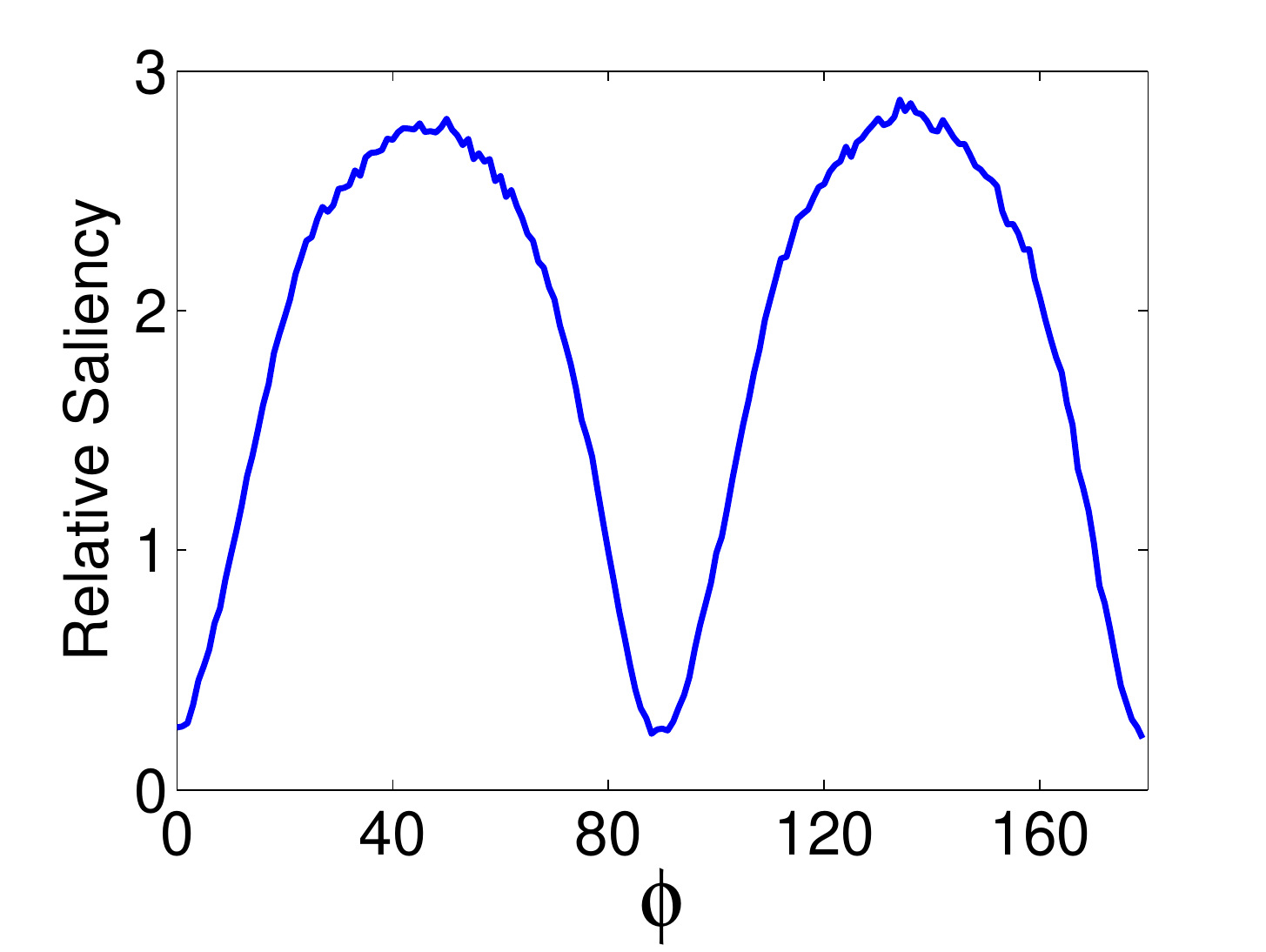}}
  \centerline{(c)}\medskip
\end{minipage}
\vspace{-0.5cm}
\caption{Illustration of the method used to predict relative saliency of the ROI}
\label{fig:OurModel}
\end{figure}

A model proposed in~\cite{Mateescu2013}, which estimates the relative changes in saliency of a ROI under rotation, demonstrates what can be accomplished by avoiding this inefficiency. Upon selecting a ROI, e.g., the tile in Fig.~\ref{fig:OurModel}(b), edge distributions within this ROI and its surroundings are obtained using statistical Hough Transform.
The solid red curve in Fig.~\ref{fig:OurModel}(a) shows that the ROI in Fig.~\ref{fig:OurModel}(b) primarily consists of horizontal ($\theta = 0^{\circ}/180^{\circ}$) and vertical ($\theta = 90^{\circ}$) edges, with a similar composition for the surrounding region, shown by the blue curve. If we were to rotate the ROI by $45^{\circ}$, then the edges of the tile would be primarily oriented at $45^{\circ}$ and $135^{\circ}$. Therefore, a counter-clockwise rotation of the ROI can be represented as a leftward circular shift of its corresponding distribution, as shown in Fig.~\ref{fig:OurModel}(a) by the dashed red curve.

\begin{figure}[!h]
\begin{minipage}[b]{.48\linewidth}
  \centering
  \centerline{\includegraphics[width=7.5cm]{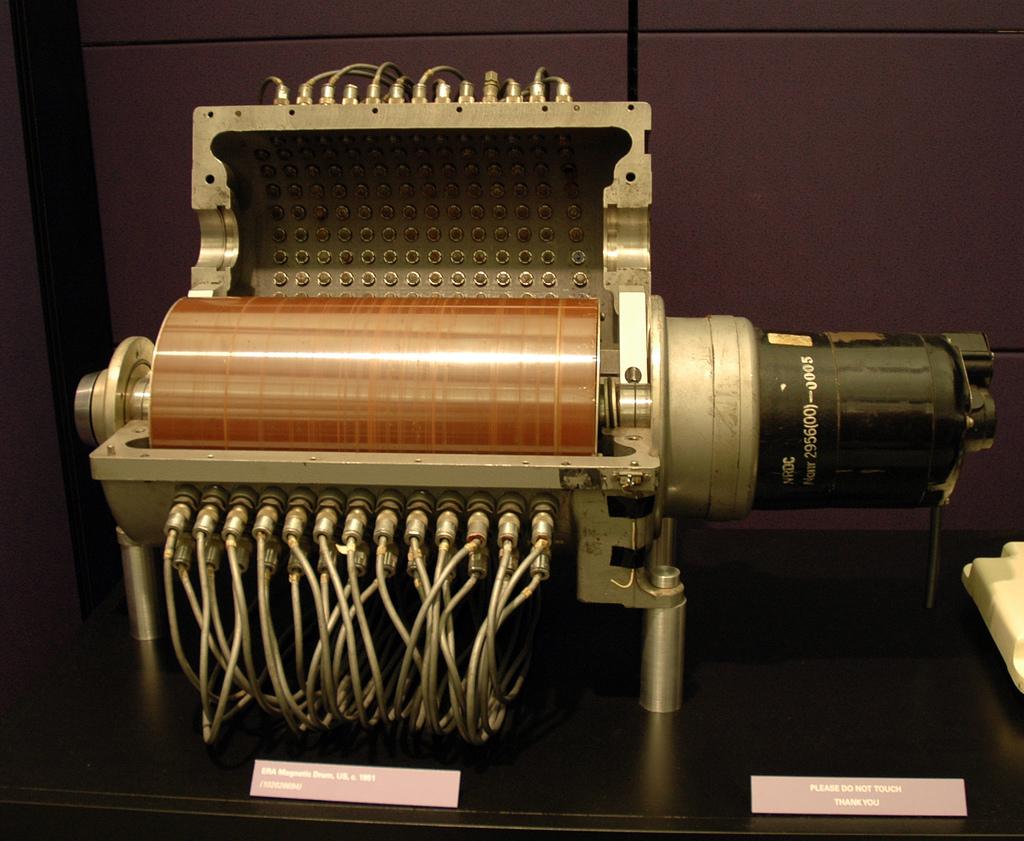}}
  \centerline{(a)}\medskip
\end{minipage}
\begin{minipage}[b]{.48\linewidth}
  \centering
  \centerline{\includegraphics[width=7.5cm]{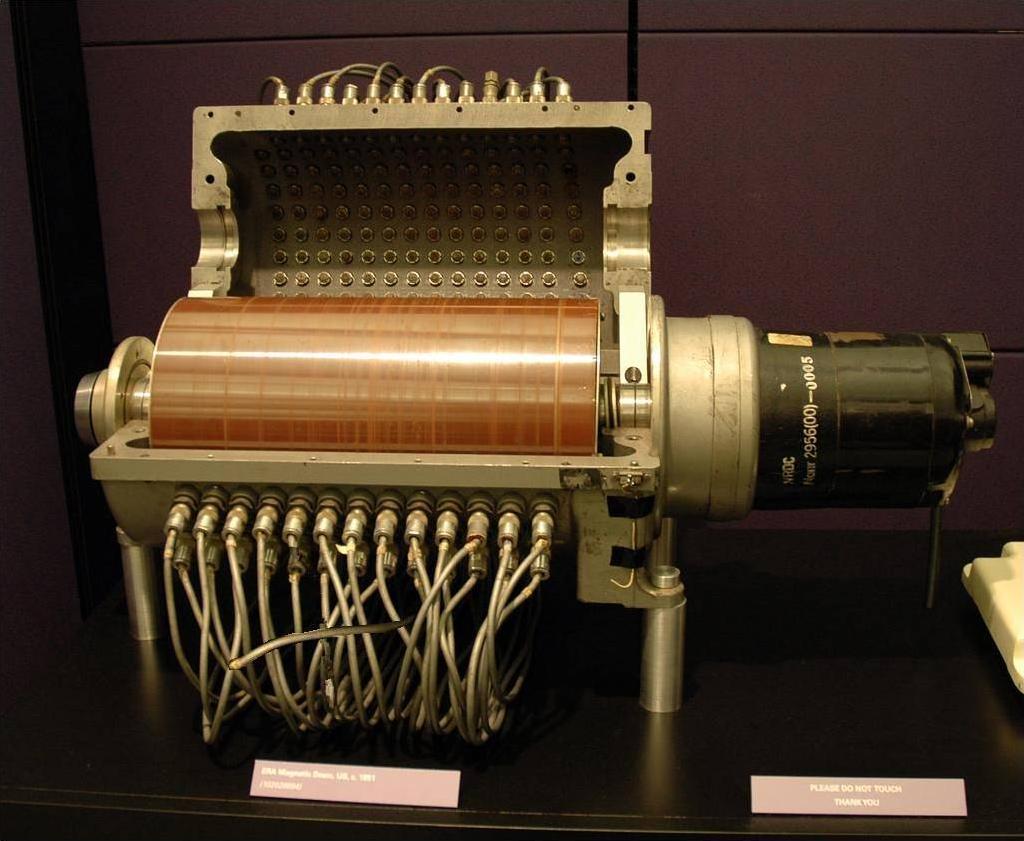}}
  \centerline{(b)}\medskip
\end{minipage}
\begin{minipage}[b]{.48\linewidth}
  \centering
  \centerline{\includegraphics[width=7.5cm]{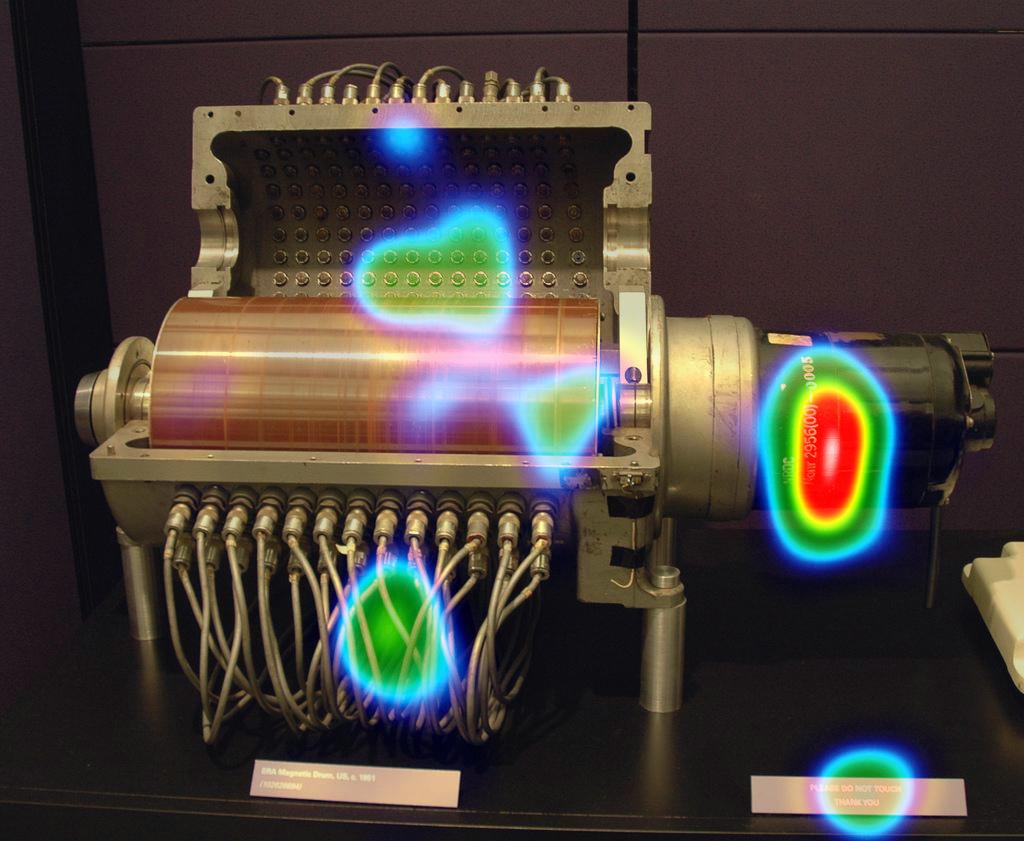}}
  \centerline{(c)}\medskip
\end{minipage}
\hspace{0.32cm}
\begin{minipage}[b]{.48\linewidth}
  \centering
  \centerline{\includegraphics[width=7.5cm]{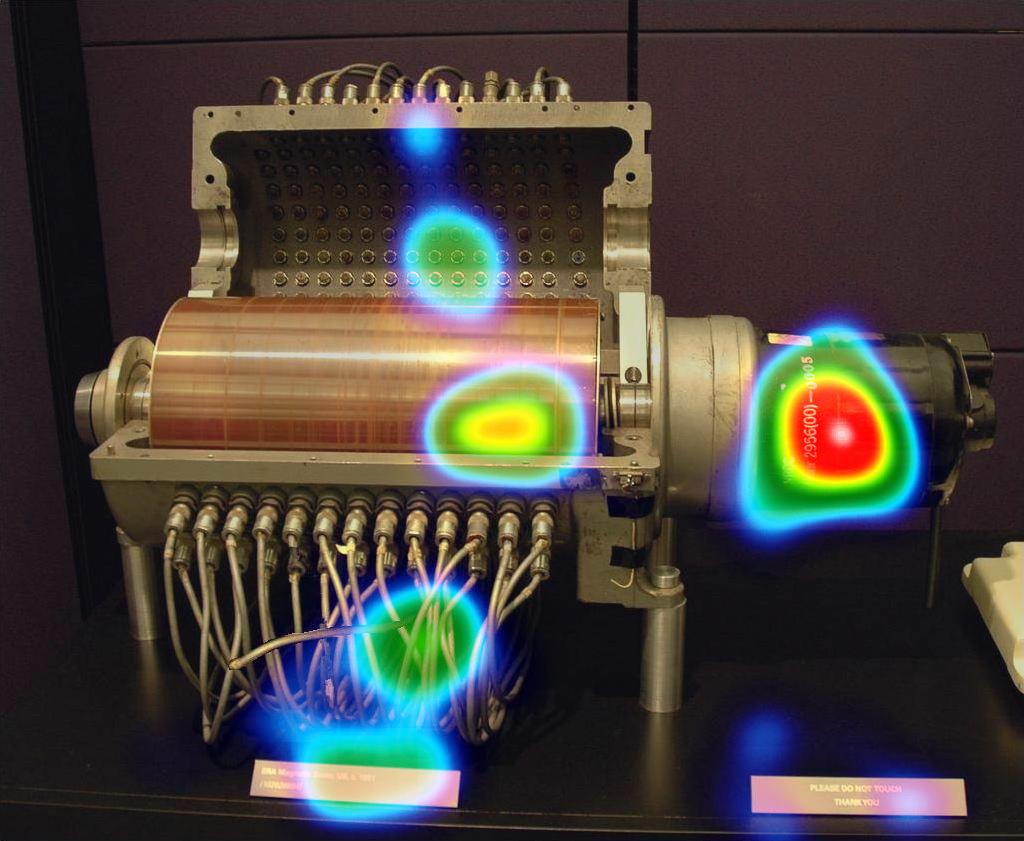}}
  \centerline{(d)}\medskip
\end{minipage}
\caption{Example of attention retargeting by orientation change~\cite{Mateescu2013} in action: (a) original image; (b) altered image - one of the cables in the bottom left has been rotated to increase saliency; (c) human gaze heat map for the original image; (d) human gaze heat map for the modified image - more attention paid to the bottom left }
\label{fig:example_orientation}
\end{figure}

The dissimilarity between the ROI and its surroundings, i.e., the center-surround operation that estimates the saliency of the ROI, is measured in~\cite{Mateescu2013} as the symmetric Kullack-Leibler divergence between their respective edge distributions. We can rapidly obtain this measure for all possible rotation angles $\phi$ of the ROI due to the concise representation of this modification as a circular shift of a 1-D distribution. The result, shown in Fig.~\ref{fig:OurModel}(c), is a complete inverse mapping of the ROI's relative saliency to rotation angle (note that a one-to-one mapping between saliency and images still does not exist, despite only a single feature being modified). The approach taken here chooses features in a manner that allows a simpler representation of the modification being made, and utilizes a center-surround operation that avoids redundancy by only focusing on the ROI. These two aspects grant the ability to rapidly predict the outcome of all possible modifications of the particular feature, in this case edge orientation. Another example of the orientation change by the method from~\cite{Mateescu2013} is shown in Fig.~\ref{fig:example_orientation}. A similar approach has been applied to hue in~\cite{Mateescu2014Color} for color-based attention retargeting.

Aside from benefits to computational complexity, this simplification may also allow more freedom in the problem formulation, particularly in the objective function used in the optimization. For example, a recently proposed model by Nguyen \textit{et al.} used a graph-based optimization for attention retargeting~\cite{Nguyen2013}. The input image is segmented into patches, and the set of patches $i$ comprising the ROI undergo color transfer from a set of candidate patches $x_{i}$. The candidate patches are mined from a large image dataset and correspond to the same objects found in the ROI so that color transfer only occurs between similar objects in an effort to maintain naturalness. Candidate patches for color transfer to the ROI are found by minimizing
\begin{equation}
E(x) = \sum\limits_{i\;\in\;\text{ROI}} E_{d}(x_{i}) + \lambda\sum\limits_{i}\sum\limits_{j\;\in\;N(i)} E_{s}(x_{i},x_{j}).
\label{eq:NguyenEnergy}
\end{equation}

The data cost $E_{d}$ is designed to consider global center-surround differences, rather than local ones. It demands that the ideal candidate patch be highly dissimilar from the entire image content outside of the ROI.  The smoothness energy $E_{s}$ is designed to punish dissimilarity between neighboring patches that are of a similar color, e.g., belonging to the same object, and encourage it otherwise. In contrast to the data cost $E_{d}$, this considers local center-surround differences to further enhance saliency. At the same time, it ensures that different objects in the ROI remain individually consistent in appearance.

\subsection{Summary}
\label{subsec:summary}

A summary of attention retargeting methods---what they aim to achieve, their methodology, and any constraints imposed---is presented in Table~\ref{tab:summary}. Most existing methods are concerned with guiding attention towards a region by increasing its saliency~\cite{Mateescu2014Color, Wong2011, Hagiwara2011, Kim2006, Kim2008, Mateescu2013, Nguyen2013}. The problem of repelling attention by reducing saliency is less explored (\cite{Su2005}).
From the analysis above, it is reasonable to conclude that~\cite{Wong2011} and~\cite{Hagiwara2011} are the most computationally demanding since they use full-scale saliency computation. Although setting hard limits on the modifications may be the simplest way to preserve aesthetics (\cite{Wong2011, Su2005, Kim2006, Kim2008} and to a lesser extent~\cite{Su2005}), they may not match the versatility of adaptive constraints~\cite{Nguyen2013}. Since an objective comparison of these methods currently does not exist, we will refrain from speculating any further on their strengths and weaknesses.

\begin{table}
	\caption{Summary of attention retargeting methods.}
	\fontsize{8}{10}\selectfont 
	\centering
    \begin{tabular}{ | P{0.08\linewidth} | p{0.15\linewidth} | p{0.35\linewidth} | p{0.30\linewidth} |}
    \hline
    \textbf{Method} & \textbf{Goal} & \textbf{Features and Approach} & \textbf{Aesthetics Constraints} \\ \hline 
    Wong and Low~\cite{Wong2011} & Change saliency of each image segment according to specified order of importance. & 
		\begin{itemize*}
			\item Intensity, chromaticity, sharpness.
			\item Iterative adjustment of each segment of the image.
			\item Uses saliency computation as a black box.
		\end{itemize*} &
		Upper and lower limits for the amount that each segment can be changed with respect to each feature. \\ \hline
    Hagiwara \textit{et al.}~\cite{Hagiwara2011} & Make ROI the most salient part of the image. & 
		\begin{itemize*}
			\item Intensity and color.
			\item Iterative adjustment of each pixel in the entire image.
			\item Uses feedback from saliency computation.
		\end{itemize*} &
		None. \\ \hline
    Su \textit{et al.}~\cite{Su2005} & Decrease saliency of textures in ROI. & 
		\begin{itemize*}
			\item Spatial frequency.
			\item Image is decomposed with steerable pyramids.
			\item Saliency subtraction is mapped directly as a scaling of steerable coefficients.
		\end{itemize*} &
		\begin{itemize*}
			\item Scaling factors with extreme values are clamped.
			\item Histogram of reconstructed output image is matched to that of the original input.
		\end{itemize*} \\ \hline
		Kim and Varshney \cite{Kim2006, Kim2008} & Increase the relative saliency of the ROI. & 
		\begin{itemize*}
			\item Intensity, chromaticity~\cite{Kim2006} and curvature~\cite{Kim2008}.
			\item Express saliency as a matrix multiplication; invert the operation with a binary map of the ROI.
			\item Scale features at each pixel/voxel within the ROI and local surroundings.
		\end{itemize*} &
		Scaling factors are normalized to a certain range. \\ \hline
		Mateescu and Baji\'{c}~\cite{Mateescu2014Color, Mateescu2013} & Map relative ROI saliency to all possible modifications of the ROI. & 
		\begin{itemize*}
			\item Hue in~\cite{Mateescu2014Color}, edge orientation in~\cite{Mateescu2013}.
			\item Efficiently represent the modification.
			\item Measure dissimilarity between ROI and surround for all possible modifications.
		\end{itemize*} &
		\begin{itemize*}
			\item No direct aesthetics constraints.
			\item Shows which modifications have a lighter impact on saliency, and hence more subtle.
		\end{itemize*} \\ \hline
		Nguyen \textit{et al.}~\cite{Nguyen2013} & Increase the relative saliency of the ROI. & 
		\begin{itemize*}
			\item Intensity and color.
			\item Graph-based optimization that maximizes global and local dissimilarity of ROI by changing the intensity and color of each object within.
		\end{itemize*} &
		\begin{itemize*}
			\item New color of objects in ROI can only come from similar objects in an image dataset.
			\item Graph-based optimization maintains the color-consistency of each object in ROI.
		\end{itemize*} \\
    \hline
    \end{tabular}
		\label{tab:summary}
\end{table}

\section{Subliminal Attention Guiding}
\label{sec:Subliminal}

Subtlety is an important aspect of attention retargeting. Modifications that degrade an image so severely that they detract from the viewing experience are of limited practical use, regardless of whether they guide viewers' attention as intended. It would be of great value to be able to guide attention in the least intrusive manner possible. Ideally, the modifications wouldn't be perceivable at all, thus guiding attention \textit{subliminally}. Retargeting attention without any perceivable alteration to the original stimuli may seem like an implausible concept. However, it is a topic of research in neuroscience and psychology that has received a fair bit of interest.

Experiments on subliminal orienting of attention generally consist of reaction-time target detection tasks. In a typical experiment, each trial begins with the participant fixating at the center of a blank screen. After some time, a subliminal cue, i.e., a visually unperceivable stimulus (experimentally confirmed to be unperceivable), is presented on the screen at a random location from a set of predetermined possible locations. Sometime later, a target, i.e., a visible stimulus, is presented randomly at one of these locations. Prior to the experiment, the participant is told to respond as quickly as possible to the target by issuing a key-press corresponding to the target's location. If the subliminal cue is truly capable of orienting the viewer's attention, reaction times to the target are expected to be shorter in trials where the cue is presented in the same location as the target. In this situation, the cue would serve as a subconscious hint to the location of the upcoming target. However, reaction times are expected to increase in trials where the cue and the target appear in different locations. Here, the cue would subconsciously distract the viewer.

\begin{figure}
	\centering
	\centerline{\includegraphics[width=15cm]{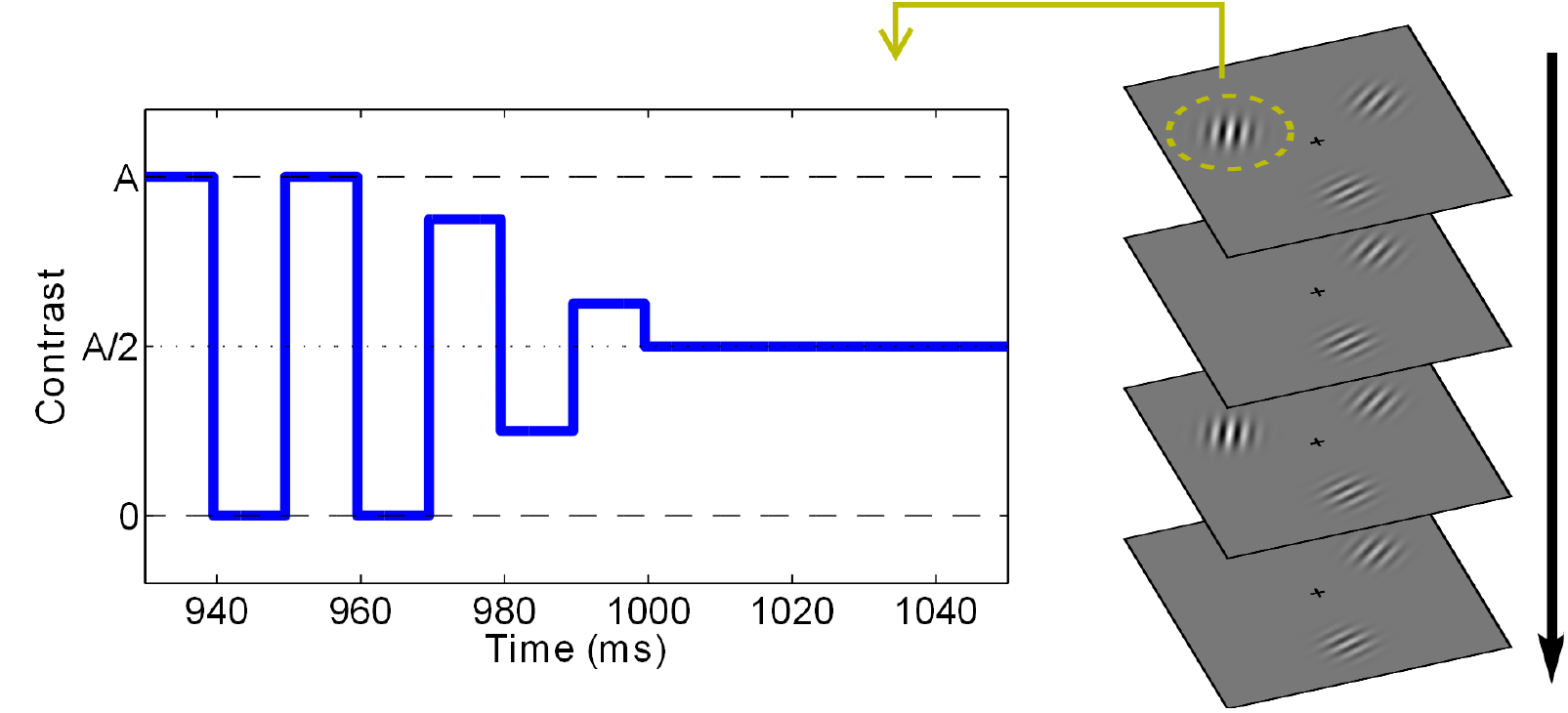}}
\captionsetup{justification=centering}
\caption{Experimental setup used in~\cite{Cheadle2011} to investigate the effects of subliminal flicker on attention.}
\label{fig:CheadleExperiment}
\end{figure}

A study by Cheadle \textit{et al.} investigates subliminal flicker as a cue for orienting attention~\cite{Cheadle2011}. Their display consists of three Gabor patterns, equally spaced on an invisible circle around the black fixation cross at the center, as shown in Fig.~\ref{fig:CheadleExperiment}. The subliminal cue in their experiments consists of alternating the contrast of one of the three Gabor patches from maximum to minimum at a frequency of 50 Hz (corresponding to a 100 Hz refresh rate). The flicker is no longer discernable at such a high frequency and the flickering Gabor patch appears identical to the other two. The target that follows the subliminal cue is a change in spatial frequency of one of the three Gabor patches. Participants' reactions to the target were found to be, on average, 15 ms faster in cases where the cue is presented in the same location as the target, which suggests that subliminal flicker is capable of drawing attention.


A temporally asynchronous cue was shown to subliminally draw attention by Mulckhuyse \textit{et al.}~\cite{Mulckhuyse2007}. In their experiment, 
they displayed three grey circular discs arranged horizontally, one of which, either the left or the right, appeared a short instant (16 ms) before the other two. In their study, they determined that participants could not accurately locate the disc that appeared earlier, suggesting that all three discs were perceived to appear simultaneously. The target consisted of a black dot that appeared inside either the left or right circular discs. Participants were told to locate the target as fast as possible. Reaction times were 11 ms shorter, on average, in cases where the target was in the same location as the cue.

Although these studies demonstrate that it may be possible to guide attention without awareness, they all use synthetic stimuli, and they used reaction time as a proxy for attention. It is uncertain whether such effects on attention would be observed if similar cues were to be applied in natural images and videos. Since reaction-time target detection tasks are less suitable for natural stimuli, investigation of subliminal orienting in natural images and videos would require eye-tracking as a more direct measure of visual attention. One of the two existing studies to have done this is the work by Huang \textit{et al.}~\cite{Huang2012}, who displayed a circular blob (radius of 50 pixels and Weber contrast of 20\%) for 50 ms prior to displaying an image in an attempt to subconsciously draw attention toward the blob's location. 
The other study~\cite{Mateescu2014Flicker} investigated the effect of subliminal flicker in natural images, where flicker was introduced to selected regions alternating the contrast in these regions from high to low at a frequency of 50 Hz. A comparison of eye-tracking data between participants who viewed the flickering images against those who viewed the original images suggests that subliminal flicker may, on average, repel attention rather than attract it~\cite{Mateescu2014Flicker}. However, a more detailed analysis is required to assess statistical significance of these findings.\footnote{A more detailed study of this effect, subsequently termed \textit{flicker observer effect}, has been presented in~\cite{FOE_2017}.}

\section{Conclusions and Future Work}
\label{sec:Conclusion}

As a relatively unexplored topic, attention retargeting presents a wide array of possibilities for improvement. For measurable progress, it is necessary to establish an objective framework for comparison of various approaches. Changing saliency is easy, but doing so in an aesthetically responsible manner is not. A reliable measure of aesthetics is of critical importance in this regard. Reverse-engineering existing saliency models may provide valuable insight and could pave the way to tractable models for predicting how simultaneous changes in multiple features affect saliency, which would be an important step forward. Subliminal attention retargeting may be considered an ultimate goal of this research direction, but further work is needed to investigate its plausibility in natural images and video. We hope that this article sparks interest in attention retargeting and motivates others to contribute.

\bibliographystyle{IEEEtran}
\bibliography{refs}

\begin{thebibliography}{10}
\providecommand{\url}[1]{#1}
\csname url@samestyle\endcsname
\providecommand{\newblock}{\relax}
\providecommand{\bibinfo}[2]{#2}
\providecommand{\BIBentrySTDinterwordspacing}{\spaceskip=0pt\relax}
\providecommand{\BIBentryALTinterwordstretchfactor}{4}
\providecommand{\BIBentryALTinterwordspacing}{\spaceskip=\fontdimen2\font plus
\BIBentryALTinterwordstretchfactor\fontdimen3\font minus
  \fontdimen4\font\relax}
\providecommand{\BIBforeignlanguage}[2]{{%
\expandafter\ifx\csname l@#1\endcsname\relax
\typeout{** WARNING: IEEEtran.bst: No hyphenation pattern has been}%
\typeout{** loaded for the language `#1'. Using the pattern for}%
\typeout{** the default language instead.}%
\else
\language=\csname l@#1\endcsname
\fi
#2}}
\providecommand{\BIBdecl}{\relax}
\BIBdecl

\bibitem{Borji_Itti_PAMI_2013}
A.~Borji and L.~Itti, ``State-of-the-art in visual attention modeling,''
  \emph{IEEE Trans. Pattern Anal. Mach. Intell.}, vol.~35, no.~1, pp. 185--207,
  Jan. 2013.

\bibitem{Tsotsos1990}
J.~K. Tsotsos, ``Analyzing vision at the complexity level,'' \emph{Brain Behav.
  Sci.}, vol.~13, pp. 423--469, 2001.

\bibitem{Itti2001}
L.~Itti and C.~Koch, ``Computational modeling of visual attention,''
  \emph{Nature Rev. Neurosci.}, vol.~2, no.~3, pp. 194--203, Mar. 2001.

\bibitem{Treisman1980}
A.~M. Treisman and G.~Gelade, ``A feature-integration theory of attention,''
  \emph{Cogn. Psychol}, vol.~12, pp. 97--136, 1980.

\bibitem{Mateescu2014Color}
V.~A. Mateescu and I.~V. Baji\'{c}, ``Attention retargeting by color
  manipulation in images,'' in \emph{Proc. ACM PIVP'14}, Orlando, FL, Nov.
  2014, pp. 15--20.

\bibitem{Wolfe2004}
J.~M. Wolfe and T.~S. Horowitz, ``What attributes guide the deployment of
  visual attention and how do they do it?'' \emph{Nature Rev. Neurosci.},
  vol.~5, pp. 1--7, 2004.

\bibitem{Itti1998}
L.~Itti, C.~Koch, and E.~Niebur, ``A model of saliency-based visual attention
  for rapid scene analysis,'' \emph{IEEE Trans. Pattern Anal. Mach. Intell.,},
  vol.~20, no.~11, pp. 1254--1259, Nov. 1998.

\bibitem{Wong2011}
L.~Wong and K.~Low, ``Saliency retargeting: An approach to enhance image
  aesthetics,'' in \emph{IEEE Workshop on Applicat. of Comput. Vision
  (WACV'11)}, Kona, HI, Jan. 2011, pp. 73--80.

\bibitem{Hagiwara2011}
A.~Hagiwara, A.~Sugimoto, and K.~Kawamoto, ``Saliency-based image editing for
  guiding visual attention,'' in \emph{Proc. PETMEI'11}, Sep. 2011, pp. 43--48.

\bibitem{Su2005}
S.~L. Su, F.~Durand, and M.~Agrawala, ``De-emphasis of distracting image
  regions using texture power maps,'' in \emph{Proc. Fourth IEEE Int'l Workshop
  Texture Analysis and Synthesis (Texture '05)}, 2005, pp. 119--124.

\bibitem{Kim2006}
Y.~Kim and A.~Varshney, ``Saliency-guided enhancement for volume
  visualization,'' \emph{IEEE Trans. Vis. Comput. Graphics}, vol.~12, no.~5,
  pp. 925--932, Sep. 2006.

\bibitem{Kim2008}
------, ``Persuading visual attention through geometry,'' \emph{IEEE Trans.
  Vis. Comput. Graphics}, vol.~14, no.~4, pp. 772--782, Jul.-Aug. 2008.

\bibitem{Mateescu2013}
V.~A. Mateescu and I.~V. Baji\'{c}, ``Guiding visual attention by manipulating
  orientation in images,'' in \emph{Proc. IEEE ICME'13}, Jul. 2013.

\bibitem{Nguyen2013}
T.~V. Nguyen, B.~Ni, H.~Liu, W.~Xia, J.~Luo, M.~Kankanhalli, and S.~Yan,
  ``Image re-attentionizing,'' \emph{IEEE Trans. Multimedia}, vol.~15, no.~8,
  pp. 1910--1919, Dec. 2013.

\bibitem{Cheadle2011}
S.~W. Cheadle, A.~Parton, H.~J. M{\"u}ller, and M.~Usher, ``Subliminal gamma
  flicker draws attention even in the absence of transition-flash cues,''
  \emph{J. Neurophysiol.}, vol. 105, no.~2, pp. 827--833, Feb. 2011.

\bibitem{Mulckhuyse2007}
M.~Mulckhuyse, D.~Talsma, and J.~Theeuwes, ``Grabbing attention without
  knowing: Automatic capture of attention by subliminal spatial cues,''
  \emph{Visual Cognition}, vol.~15, no.~7, pp. 779--788, 2007.

\bibitem{Huang2012}
T.~H. Huang, Y.~H. Yang, H.~I. Liao, S.~L. Yeh, and H.~H. Chen, ``Directing
  visual attention by subliminal cues,'' in \emph{Proc. IEEE ICIP'12}, Oct.
  2012, pp. 1081--1084.

\bibitem{Mateescu2014Flicker}
V.~A. Mateescu and I.~V. Baji\'{c}, ``Can subliminal flicker guide attention in
  natural images?'' in \emph{Proc. ACM PIVP'14}, Orlando, FL, Nov. 2014, pp.
  33--34.

\bibitem{FOE_2017}
N.~Waldin, M.~Waldner, and I.~Viola, ``Flicker observer effect: Guiding
  attention through high frequency flicker in images,'' \emph{Computer Graphics
  Forum}, vol.~36, no.~2, pp. 467--476, 2017.

\end{thebibliography}

\end{document}